\pdfoutput=1
\documentclass[10pt, a4paper]{article}
\usepackage{CJKutf8}
\usepackage{lrec}
\usepackage{multibib}
\newcites{languageresource}{Language Resources}
\usepackage{graphicx}
\usepackage{tabularx}
\usepackage{soul}
\usepackage{epstopdf}
\usepackage{amsmath}
\usepackage{booktabs}

\usepackage[group-separator={,},group-minimum-digits=3]{siunitx}

\usepackage{hyperref}

\usepackage{xstring}
\usepackage{tikz-dependency}

\newcommand*{\affmark}[1][*]{\textsuperscript{#1}}

\newcommand{\chn}[1]{\begin{CJK*}{UTF8}{gbsn}#1\end{CJK*}}
\newcommand{\chnq}[1]{\begin{CJK*}{UTF8}{gbsn}``#1''\end{CJK*}}
\newcommand{\tri}[1]{$\langle \text{#1} \rangle$}
\newcommand{\bq}{\`\ }

\makeatletter
\def\new@fontshape{}
\makeatother
\usepackage{gb4e}
\exewidth{(\thexnumi)}
\noautomath

\title{Building an Ellipsis-aware Chinese Dependency Treebank for Web Text}

\name{Xuancheng Ren\affmark[1,2], Xu Sun\affmark[1,2], Ji Wen\affmark[1,2], Bingzhen Wei\affmark[1,2], Weidong Zhan\affmark[1,3], Zhiyuan Zhang\affmark[1,2]}

\address{\affmark[1]MOE Key Laboratory of Computational Linguistics, Peking University\\
         \affmark[2]School of Electronics Engineering and Computer Science, Peking University\\
         \affmark[3]Department of Chinese Language and Literature, Peking University\\
         Beijing, China\\
         \{renxc, xusun, wenjics, weibz, zwd, zzy1210\}@pku.edu.cn\\}

\abstract{
Web 2.0 has brought with it numerous user-produced data revealing one's thoughts, experiences, and knowledge, which are a great source for many tasks, such as information extraction, and knowledge base construction. However, the colloquial nature of the texts poses new challenges for current natural language processing techniques, which are more adapt to the formal form of the language. Ellipsis is a common linguistic phenomenon that some words are left out as they are understood from the context, especially in oral utterance, hindering the improvement of dependency parsing, which is of great importance for tasks relied on the meaning of the sentence. In order to promote research in this area, we are releasing a Chinese dependency treebank of 319 weibos, containing 572 sentences with omissions restored and contexts reserved\thanks{The treebank is available at \url{https://github.com/lancopku/Chinese-Dependency-Treebank-with-Ellipsis}.}. \\ \newline  }

\begin{document}

\maketitleabstract

\section{Introduction}

With the rapid development of Web 2.0, the internet provides us with numerous raw text, which is a great source for structured learning such as information extraction. 
However, the information in Web 2.0 is usually highly unstructured and of various forms, written in a more casual way, which makes it hard to directly apply techniques that works for regular text. 
One probable way to structure the text is through syntax parsing, which exposes the syntax relation, and implicitly semantic relation, between words, especially dependency parsing \cite{tesniere1965elements}. 
However, due to the different construction of the sentences in web text, the current parsing methods face greatly challenge \cite{kubler2009dependency,Petrov12}. 
Web text usually represents more of a spoken language aspect than the written language aspect, which means the social effect is more dominant in its formation, and the efficiency in conveying the information is more valued. In this work, we focus on building a resource to facilitate the parsing of the web text, specifically text from online microblogs.

A common issue of the current techniques is that they lack the ability to deal with omitted elements in the text, which are more often referred as ellipsis in linguistics. 
Ellipsis is a common linguistic phenomenon across languages, which facilitates the communication in real-world conversations, as the omitted elements should be understood from the context. 
The down side is that the sentence is often made less structured, and cannot easily be understood if extracted from the context.
This work is a first step towards establishing a more robust parsing framework through recognizing the ubiquitous ellipsis in web text and identifying the omitted elements from such type of sentences.
We build a Chinese dependency treebank based on the messages extracted from Weibo\footnote{Weibo is China's most popular microblogging service, in comparison with Twitter. In the following, we will refer to the platform as Weibo, and each message the user posts as a weibo, just like a tweet.}, in which we restore the omitted elements, and hope it could help advance the study in this area and improve the peformance of dependency parsing of web text.

To our knowledge, this is the first study that seeks to build a treebank with focus on ellipsis in context for Chinese. 
Chinese Treebank \citelanguageresource{ctb5}, which is initially a constituent treebank, and then converted to a dependency treebank \cite{MarneffeMM06}, incorporates the idea of empty category from the government and binding theory of \newcite{chomsky1993lectures}, but is fundamentally different from ours, because empty categories usually cannot appear in a legal sentence, while ellipsis means the omitted elements can always be restored, and are omitted just for the convenience of communication.
In practice, the empty category annotation is often being ignored when building a parser, making the annotation more or less useless \cite{Campbell04}. 
Universal Treebanks \cite{McDonaldNQGDGHPZTBCL13,MarneffeDSHGNM14,NivreMGGHMMPPST16}, considering the ellipsis as a less important phenomenon, deal with it by promoting the omitted words' dependents, or use the special \textit{orphan} relation, which either produces a confusing dependency, or isolates the relation between the part with ellipsis and the part without ellipsis. This solution makes it harder for the parser to learn from the data.

It is also necessary to point out the difference from the study of pronoun-dropping, where certain classes of pronouns may be omitted when they are inferable. For example, in the sentence \chnq{谁说的？}(Who say \textit{DE\textsubscript{particle}}, i.e., ``Who said it?''), the object of \chnq{说}(``say''), that is, ``it'' in the translation, is omitted or dropped.
There is a huge amount of research regarding Chinese from a theoretical linguistic view. The concept is overlapped with the concept of ellipsis, more specifically contained in the concept of ellipsis. While in pronoun-dropping the omitted elements is supposed to be pronouns or noun phrases, the constraint does not apply to ellipsis. In spoken Chinese, verb phrases can also be omitted. To our knowledge, pronoun-dropping is also not applied or reflected when building treebanks.

In what follows, we will start by explaining ellipsis in Chinese Language, then briefly review the main considerations and the main steps we take in order to produce consistent and helpful annotations for the related natural language processing tasks, and finally introduce the annotated dataset with basic analysis of its attributes.

\section{Ellipsis}
To annotate ellipsis in sentences, we first need to define what should be considered as ellipsis. In linguistic theory, the specification of ellipsis has been a unsolved issue for a long time. In traditional natural language processing, ellipsis is not recognized as an important factor, which may be a measure of expediency at the time, as the goal is to process the regular form first.
That no longer holds with the rapid adoption of Web 2.0, characterized by the publicness, informality, and the causal expression.

Following the theories in Chinese linguistics \cite{chao1965grammar}, we try to give a definition of ellipsis in the practice of natural language processing, whose goal is to assist the dependency parsing, which reveals the syntactic role and, to some extent, semantic role of the word in a sentence.

\subsection{Ellipsis in Chinese Language}

We define ellipsis in Chinese as textual omission of syntactic components, specifically words or phrases, expressing a semantic role in a sentence, that are optional but not obligatory in an utterance, and if elided, given the context of the sentence, the exact wording or, if referring to a object or concept, at least a board category of what the ellipsis refers to, shall be determined. There are four main parts in the definition, which are:
\begin{itemize}
\item It happens at the level of words and phrases. The omission of characters or morphemes is not considered ellipsis.
\item The elided words must express meaning related to the sentence. If the words are not helpful to the understanding of the sentence, the omission is ignored.
\item The elided words can be said. If they are said but the resulting sentence becoming incorrect or illegal, that kind of omission is not what ellipsis considers.
\item The elided words can be determined from the context. If we do not know what the elided refers to or stands for, we do not treat the omission as ellipsis.
\end{itemize}

The definition is the guiding principle in our annotation of the text. It is worth noticing that the definition is not from a pure syntax view. As the goal of the annotation is to broaden the use of the web text, the semantic side is paid more attention to.

\section{Construction of the Dataset}

Following the definition, we develop several considerations in the construction of the dataset, which we will introduce in Section \ref{sec:considerations}, then we will show our annotation procedure in Section \ref{sec:procedure}, and finally we will explain the format of the dataset in Section \ref{sec:format}.

\subsection{Considerations\label{sec:considerations}}

\textbf{Speakability} 
The omission can be said in a regular sentence, which means the omission is optional rather than obligatory. 
For example, in the well-known example, \chnq{我请他吃饭。}(I invite him eat meal, i.e., ``I invite him to eat a meal.''), \chnq{他}(``him'') is both the object of \chnq{请}(``invite'') and the subject of \chnq{吃}(``eat''), which causes an illegal situation where two dependency point at \chnq{他}(``him''). It's common belief that a \chnq{他}(``him'' or ``he'') is dropped from the sentence\footnote{However, it has not reached consensus whether the subject is dropped or the object is dropped.}. 
However, if restored, the resulting sentence \chnq{我请他他吃饭。}(I invite him he eat meal, i.e., ``I invite him to eat a meal.'') is unspeakable for a Chinese speaker. 
Although the omitted word play a syntactic and semantic role in the sentence, due to the speakablility, we do not restore the dropped elements\footnote{This is called PRO-drop from the view of empty category. \cite{huang1989pro}}.
\medskip

\textbf{Identifiability} 
The omission must be known from the context. 
In traditional Chinese grammar, this constraint is so strong that it requires the omission can be restored uniquely and unambiguously.
However, in practice, to make a single sentence semantically reasonable, we are aware that some words are missing, and we know vaguely what they refer to, but we cannot restore the words uniquely. 
For example, in the sentence \chnq{这样就没了。} 
(This way just not-exist \textit{LE\textsubscript{particle}}, i.e., ``It has gone this way.''), 
the subject of \chnq{没}(not-exist, i.e., ``vanish'') is omitted, which causes the problem that most parser will treat \chnq{这样}(this way, i.e., ``this way'') as the subject. 
However, most of the time, we don't know what the exact wording of the omission is, but we are aware that it must be a noun and represent a thing. 
Due to the semantic importance, we introduce several categories in this case, that is THG, EVT, PPL, and OTH, representing concrete or abstract things, events, people-like perceivers, and others respectively.
\medskip

\textbf{Necessity} 
If the omission does not affect the syntactic or semantic side of the sentence, we just ignore it and do not restore the omission. 
For example, in the sentence \chnq{我喜欢看书，但她不喜欢。} 
(I like read book, but she not like, i.e., ``I like reading books, but she doesn't.''
\footnote{The sentence is indeed an elliptical construction. 
But the formation is different between Chinese and English. 
The Chinese one elides the object of \chn{``喜欢''}(``like''), but keeps the predicate, while the English one omits both the predicate and its object, which is particularly frequent in English, commonly known as verb phrase ellipsis or VPE. 
}), 
\chnq{但}(``but''), which is an adversative conjunction, is a short form of \chnq{但是}(but be, i.e, ``but'') with the character \chnq{是}(be) omitted. 
However, the character \chnq{是}(be) here does not affect the syntax or meaning of the sentence. 
Hence, we do not restore \chnq{是}(be) here, although in a more formal utterance, \chnq{是}(be) is needed.

\subsection{Annotation Procedure\label{sec:procedure}}
We will first give a brief introduction to the raw text we used, and then describe the annotation procedure we practiced.

The raw text we used is Leiden Weibo Corpus (LWC)\footnote{http://lwc.daanvanesch.nl/}, which consists of more than 5 million messages posted on Weibo in January 2012. 
The period contains normal weeks, as well as the Chinese New Year holiday, which makes the corpus includes general topics in a normal weeks and also a prominent topic, i.e. the Chinese New Year. 

The main reason that we choose the text from microblogs is that text from microblogs is more useful than the text from typical blogs for information extraction. Besides, the text from microblogs is often more oral and casual, which means the ellipsis probably occurs more. While some may argue that the ellipsis comes from the 140 characters limitation rather than the attributes of the language itself, we believe that, different from the 140 letters limitation of a tweet, a lot things can be conveyed in 140 Chinese characters of a weibo, and the character limitation is not a dominant factor to the ellipsis.

\begin{figure}[ht]
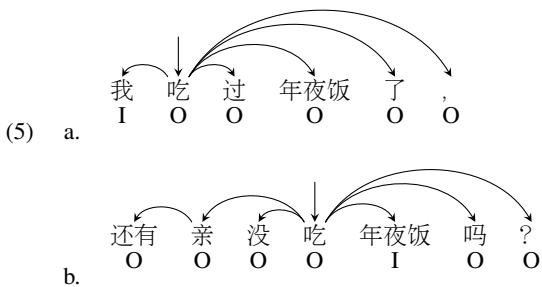

\footnotesize
\begin{exe}
    \ex \chn{吃过年夜饭了，还有亲没吃吗？}
    \ex \begin{xlist}
        \ex \chn{吃过年夜饭了，}
        \ex \chn{还有亲没吃吗？}
        \end{xlist}
    \ex \begin{xlist}
        \ex \chn{\bq我\bq吃过年夜饭了，}
        \ex \chn{还有亲没吃\bq年夜饭\bq吗？}
        \end{xlist}
    \ex \begin{xlist}
        \ex \chn{\bq我\bq/吃/过/年夜饭/了/，}
        \glt {\bq}I\bq eat \textit{HAVE\textsubscript{particle}} family-reunion-dinner \textit{LE\textsubscript{particle}} ,
        \trans ``I have eaten the family reunion dinner.''
        \ex \chn{还有/亲/没/吃/\bq年夜饭\bq/吗/？}
        \glt any-other person not eat {\bq}family-reunion-dinner\bq \textit{MA\textsubscript{particle}} ?
    	\trans ``Has any of you not eaten the family reunion dinner?''
        \end{xlist}
    \ex \begin{xlist}
        \ex \begin{dependency}[hide label, arc edge]
            \begin{deptext}[column sep=8pt]
            \chn{我} \& \chn{吃} \& \chn{过} \& \chn{年夜饭} \& \chn{了} \& \chn{，} \\
            I \& O \& O \& O \& O \& O \\
            \end{deptext}
            \depedge{2}{1}{}
            \depedge{2}{3}{}
            \depedge{2}{4}{}
            \depedge{2}{5}{}
            \depedge{2}{6}{}
            \deproot[edge unit distance=1.5ex]{2}{}
            \end{dependency}
        \ex \begin{dependency}[hide label, arc edge]
            \begin{deptext}[column sep=8pt]
            \chn{还有} \& \chn{亲} \& \chn{没} \& \chn{吃} \& \chn{年夜饭} \& \chn{吗} \& \chn{？} \\
            O \& O \& O \& O \& I \& O \& O \\
            \end{deptext}
            \depedge{2}{1}{}
            \depedge{4}{2}{}
            \depedge{4}{3}{}
            \depedge{4}{5}{}
            \depedge{4}{6}{}
            \depedge{4}{7}{}
            \deproot[edge unit distance=1.5ex]{4}{}
            \end{dependency}
        \end{xlist}
\end{exe}
\vspace{-10pt}
\caption{An example of the annotation procedure.\label{ex}}
\vspace{-5pt}
\end{figure}

The annotation is done in the following order, and an example is given in Table \ref{ex}.

\textbf{Selection}
As the corpus is much too large for our purpose, we randomly select weibos from the corpus. A major attribute of ellipsis is that it should be understood from the context. However, in LWC, a single weibo is a stored unit, lacking the relations between weibos, such as forwarding and replying. 
Hence, some omissions just cannot be restored, even if we use the aforesaid general categories. 
Besides, there are also tons of advertisements and chaos messages in the corpus. To deal with the problem, we further purge the randomly selected weibos by considering the content of the weibos. 
We only keep the weibos that are roughly about normal daily topics, and drop the rest. 
In the end, we got \num{319} weibos from \num{500} random chosen weibos. The weibos contain a total of \num{8382} tokens.
\medskip

\textbf{Sentence Split}
We then manually split the sentences in a weibo, as a weibo generally contains more than one sentences, but the sentence split is often implicit, due to the casual use of punctuations. Besides, sentences, in the sense of English or other Indo-European languages, in Chinese does not only end at the periods, but also can stop at the commas. The reason is that sentences are normally formed on a topic level, and several sentences, which are independent from each other grammatically, and lack conjunctions in between, are grouped as a sentence because they related to a certain topic.
By manually splitting the sentences, a weibo can be seen as the context of its sentences, which is necessary for omission restoration tasks.
The split standard is that if the topic or the subject is changed between the parts split by commas, we split the parts. This step gives us \num{572} sentences from \num{319} weibos. 
\medskip

\textbf{Restoration}
Restoration is the major step in our annotation, and also the step we pay the most attention to. The considerations in annotation is explained in Section \ref{sec:considerations}.  
In this step, we restored \num{208} words of \num{256} characters.
\smallskip

All of the previous steps are done in the .txt files.
\medskip

\textbf{Word Segmentation}
From this step, the annotation becomes more demanding, and we gain assistance from the annotation tool brat \cite{DBLP:conf/eacl/StenetorpPTOAT12}\footnote{http://brat.nlplab.org/} and the Stanford CoreNLP tool\footnote{https://stanfordnlp.github.io/CoreNLP/}. The word segmentation procedure basically follows the guideline of CTB \cite{xia2000segmentation}. The difference is that we treat all the words, which typically only appear in web text, as single words, regardless of the boundness of its morphemes/characters. For example, the word \chnq{给力} (give power, i.e. ``forceful'') is not segmented into \chnq{给} (``give'') and \chnq{力} (``power''). We also mark the restored words with the tag ``I'' to distinguish from the original words. The word segmentation annotation generated from Stanford CoreNLP tool are considered as baselines, which are already available in LWC. A total of \num{8590} words are found in this step, including the restored words.
\medskip

\textbf{Dependency Arc Annotation}
This is the core of any dependency treebanks, which is also the most challenging part. With limited resources, this step is simplified as much as possible. We only annotate the dependency relation between words, excluding the type of the dependency, and some words are dropped in this step, if the words are emoji converted words, cause the dependency tree to be non-projectable or are less relevant to the sentence. For example, there are a lot of interjections, or called exclamations, such as \chnq{嗯}, \chnq{哦}, and \chnq{哈} in the sentences. 
Although they are part of the sentence, and often are at the start, we do not annotate their dependencies, because the relation between them and the sentence is rather weak. 
Nonetheless, we do annotate the modal particles' dependencies, such as \chnq{吗}, \chnq{呢}, and \chnq{吧}, as they do affect the modal of the sentence. 
For example, \chnq{吗} is a distinctive mark for interrogative sentence. 
This gives us \num{8018} dependencies.

\subsection{Annotation Format\label{sec:format}}

There are two types of files after the annotation. One is a text file containing the split weibos and restored content, and the other is an annotation file, containing the word segmentation, omission restoration, and the dependency annotation.

In the text file, each line contains a sentence, and a blank line separates weibos. Each restored element is surrounded by the grave accent, i.e., the back quote mark ``\bq''. The original ``\bq'' in the file is changed to ``\tri{\bq}'', which is very rare in our annotation.

The annotation file is of the same format with the brat's default format. Each line is an annotation entry, either a word or a dependency. 

For the word annotation, the line is of format 
``\tri{wid}\textbackslash t\tri{wtag} \tri{start} \tri{end}\textbackslash t\tri{word}'', 
where \tri{wid} is the word's unqiue identifier in a file, \textbackslash t means a tab, \tri{wtag} is either ``I'', meaning a restored element, or ``O'', meaning an original element, the \tri{start} and the \tri{end} give the start offset and the offset after the end of the word from the start of the file, and \tri{word} is the original form of the word in the text. Please notice, the replaced ``\bq'' is also changed to its original form, and the restored element mark ``\bq'' is not annotated.

For the dependency annotation, the line is of format ``\tri{did}\textbackslash t\tri{dtag} Arg1:\tri{hid} Arg2:\tri{cid}'', where \tri{did} is the dependency's unqiue identifier in a file, \tri{dtag} can only be ``dep'' as we do not differentiate the dependency types, and \tri{hid} and \tri{cid} give the head and dependent word ID of the dependency.

For natural language processing usage, the annotation file is adequate for ellipsis restoration, word segmentation, and dependency parsing tasks, as the related annotations are all reflected in the file.

Furthermore, we combined the annotation files into a single file in the tsv format. There are four columns in the file. The first column is the token's index in the sentence, starting from 1. The second column is the textual form of the token. The third column indicates whether the token is a restored one. 'O' stands for original tokens, and 'I' stands for restored tokens. The fourth column is the head of the token. 0 indicates the token is the root. There is an empty line between sentences, and an extra empty line between weibos.

\section{Dataset}

We are releasing a first version of the dataset, containing \num{8590} tokens, \num{572} sentences, and \num{319} weibos (Table \ref{tab:dataset}). 
The raw text is from LWC, a weibo corpus. Unsurprisingly, due to the characteristics of microblogging, the average length of the sentences are quite short, around 15.0 tokens per sentence, comparing to 27.0 tokens per sentence in CTB5. 
We have restored \num{256} characters and \num{208} words in the dataset. 
As shown in Table \ref{tab:dataset}, ellipsis is indeed a common phenomenon in web text, which requires more attention, as \num{162} of the sentences, and \num{122} of the weibos contain ellipsis, meaning 38.24\% of the weibos involve ellipsis.

\begin{table}[ht]
    \centering
    \sisetup{table-format = 5, round-mode=places, round-precision = 0}
    \setlength{\tabcolsep}{3pt}
    \begin{tabular}{l|r|r|r|r}
    \hline
        Type & \#Token & \#Word & \#Sentence & \#Weibo  \\
    \hline\hline
        Original & \tablenum{12508} & \tablenum{8382} & \tablenum{572} & \tablenum{319} \\
        Ellipsis & \tablenum{256} & \tablenum{208} & \tablenum{162} & \tablenum{122} \\ 
    \hline
        Overall & \tablenum{12764} & \tablenum{8590} & \tablenum{572} & \tablenum{319}\\
    \hline
        Percentage (\%) & \tablenum{2.01} & \tablenum{2.42} & \tablenum{28.32} & \tablenum{38.24}  \\
    \hline
    \end{tabular}
    \caption{Statistics of the dataset. Each column representing the term that are being counted, 1: the type of the term; 2: number of tokens or characters; 3: number of words; 4: number of sentences containing the type of the term; 5: number of weibos containing the type of the term. We can see that ellipsis is indeed common in the annotated dataset, as more than half of the weibos containing ellipsis.\label{tab:dataset}}
\end{table}

\begin{table}[ht]
    \centering
    \sisetup{table-format = 4}
    \begin{tabular}{r@{ $\rightarrow$ }l|c}
    \hline
        {Head} & {Dependent} & {\#Depedency}  \\
    \hline\hline
        Original & Original & \tablenum{7762}\\
        Original & Ellipsis & \tablenum{187}\\ 
        Ellipsis & Original & \tablenum{61}\\
        Ellipsis & Ellipsis & \tablenum{8}\\
    \hline
        \multicolumn{2}{c|}{Overall} & \tablenum{8018}\\
    \hline
    \end{tabular}
    \caption{Statistics of the treebank. The first column describes the type of the elements involved in a dependency. The second column gives the count of the dependency.}
    \label{tab:treebank}
\end{table}

We further show the statistics of the annotated dependencies of the dataset in Table \ref{tab:treebank}.
There are in total \num{256} dependencies involving ellipsis.
As there are only \num{208} restored words, some restored words are linked to multiple original words. 
In addition, restored words serve as heads in \num{61} dependencies, which means, if not restored, the omission will cause dependent promotion or orphan relation just like in Universal Treebanks, harming the syntactic soundness of the dependency tree. 
To our surprise, in \num{8} dependencies, both the head and the dependent are omitted, which may indicate some of the restored omissions may be redundant.
We will further investigate the case and try to refine the guidelines, if needed, in the next version of the dataset.

\section{Efforts in Maintaining Quality}

At the beginning of the annotation, it is unsurprising to see that the annotators hardly fully agree on annotations of one weibo. 
As the annotation task involves several steps, some of which are really hard even to senior students majoring in linguistics, such as restoration and dependency parsing.
Besides, the corpus is rather informal, usually with unusual utterance and wrong characters.
In order to maintain the quality of the annotations, and the consistency across the dataset, we make the following supervision efforts.

Several Peking University students majoring in computational linguistics were chosen to conduct the annotation study. 
Each student was given an annotation guideline and ten sample weibos, which we developed gold standard annotations. 
We compared their answers with the standard annotations, pointed out the differences, then provided specialized guidance for the students and revised the guideline as needed. \cite{pyysalo2014collaborative}
The test was repeated, until the quality of the annotations of the ten samples are met. Then, the students were given real text to annotate.

Due to limited annotators, each weibo was only annotated once after the test phase. The workload is divided into units of 10 weibos. Each time, a student is given a unit to annotate. After the unit is annotated, we randomly choose one weibo to double-check. If the quality is met, the unit is considered okay. If not, the rest weibos are all checked, and we select the qualified weibos, and discard the bad ones. By doing this, we could maximize the annotation speed, and, at the same time, keep a basic and satisfying quality of the first version of the dataset.

As the restoration standard is hard to unify, we make a list of the aspects that should be considered when restoring an element. For example, if a sentence is a subject-predicate construction, the elided subject must be restored, because it is very common in oral Chinese to omit the subject, which most of the time is also the topic of the conversation, but it can cause serious problems for parsers, as the subjects in consecutive sentences may refer to different things.

In annotation, especially the word segmentation step and dependency parsing step, to alleviate the workload of the annotators, we use off-the-self tools, specifically the Stanford CoreNLP tools, to automatically generate the related references. The annotators could use the annotations generated as a baseline, and rectify the wrong dependencies as they recognize. This could further facilitate the annotation procedure, and maintain the quality of the dependency at least above the performance of the used parser, which is about 83.9 in terms of UAS on the test set of CTB5 \cite{ChenM14}. In future revisions, tools with better accuracy and faster speed may be considered \cite{SunZMTT13,SunLWL14,DyerBLMS15,Sun14,XuS16,KiperwasserG16,Sun16,meprop,zhang-sun-wan:2017:CoNLL,hydoracle}.

The reason for these relaxations is that the main focus of the treebank is the annotation of ellipsis, and the study of the effect of ellipsis on dependency parsing, so as to other semantic related tasks. We shift our annotation focus to the annotation of sentence split, omission discovery and restoration, which are less demanding than the annotation of dependency trees, after we find the annotation process is too slow.

\section{Conclusion}

In this study, we introduce a practical definition for ellipsis in Chinese. We also introduce a practical scheme for ellipsis annotation, and build an ellipsis-aware Chinese dependency treebank for web text, where the elided is restored, and the necessary context is reserved. 
The dataset contains \num{572} sentences from \num{319} weibos, including \num{208} restored omissions, forming \num{8018} dependencies.
We are releasing an initial version of the dataset, and we hope the dataset will advance the study in ellipsis restoration and the following tasks in a natural language processing pipeline.

\section{Acknowledgements}

We thank the anonymous reviewers for their insightful comments. This work was supported in part by National Natural Science Foundation of China (No. 61673028), National High Technology Research and Development Program of China (863 Program, No. 2015AA015404), and an Okawa Research Grant (2016). Email correspondence to Xu Sun.

\section{Bibliographical References}
\label{main:ref}

\bibliographystyle{lrec}
\bibliography{xample}

\section{Language Resource References}
\label{lr:ref}
\bibliographystylelanguageresource{lrec}
\bibliographylanguageresource{xample}

\end{document}